\documentclass[10pt,twocolumn,letterpaper]{article}

\usepackage{iccv}
\usepackage{times}
\usepackage{epsfig}
\usepackage{graphicx}
\usepackage{amsmath}
\usepackage{amssymb}
\usepackage{stfloats}
\usepackage{makecell}
\usepackage[breaklinks=true,bookmarks=false,colorlinks=true,linkcolor=red]{hyperref}
\iccvfinalcopy % *** Uncomment this line for the final submission

\def\iccvPaperID{****} % *** Enter the ICCV Paper ID here

\ificcvfinal\fi

\begin{document}

%%%%%%%%% TITLE
\title{CondLaneNet: a Top-to-down Lane Detection Framework Based on Conditional Convolution}

\author{Lizhe Liu$^{1}$\hspace{1cm}Xiaohao Chen$^{1}$\hspace{1cm}Siyu Zhu$^{1*}$\hspace{1cm}Ping Tan$^{12}$\\
${}^{1}$Alibaba Group\hspace{1.5cm}${}^{2}$Simon Fraser University\hspace{0cm}
}

\maketitle
\ificcvfinal\thispagestyle{empty}\fi

\def\thefootnote{*}\footnotetext{Siyu Zhu is the corresponding author.}
%%%%%%%%% ABSTRACT
\begin{abstract}
Modern deep-learning-based lane detection methods are successful in most scenarios but struggling for lane lines with complex topologies.
In this work, we propose CondLaneNet, a novel top-to-down lane detection framework that detects the lane instances first and then dynamically predicts the line shape for each instance. Aiming to resolve lane instance-level discrimination problem, we introduce a conditional lane detection strategy based on conditional convolution and row-wise formulation. Further, we design the Recurrent Instance Module(RIM) to overcome the problem of detecting lane lines with complex topologies such as dense lines and fork lines. 
Benefit from the end-to-end pipeline which requires little post-process, our method has real-time efficiency.
We extensively evaluate our method on three benchmarks of lane detection. Results show that our method achieves state-of-the-art performance on all three benchmark datasets. Moreover, our method has the coexistence of accuracy and efficiency, e.g. a 78.14 F1 score and 220 FPS on CULane. 
Our code is available at  \url{https://github.com/aliyun/conditional-lane-detection}. 

\end{abstract}

%%%%%%%%% Introduction

\section{Introduction}
Artificial intelligence technology is increasingly being used in the driving field, which is conducive to autonomous driving and the Advanced Driver Assistance System(ADAS). As a basic problem in autonomous driving, lane detection plays a vital role in applications such as vehicle real-time positioning, driving route planning, lane-keeping assist, and adaptive cruise control.

Traditional lane detection methods usually rely on hand-crafted operators to extract features~\cite{liu2010combining, zhou2010novel, hur2013multi, kim2008robust, jiang2009new, borkar2009robust, jiang2010computer, tan2014novel}, and then fit the line shape through post-processing such as Hough transform~\cite{liu2010combining, zhou2010novel} and Random Sampling Consensus (RANSAC)~\cite{kim2008robust, jiang2009new}.
However, traditional methods faile in maintaining robustness in real scene since the hand-crafted models cannot cope with the diversity of lane lines in different scenarios~\cite{neven2018towards}.
Recently, most studies about lane detection have focused on deep learning~\cite{tang2020review}. Early deep-learning-based methods detect lane lines through segmentation~\cite{pan2018spatial, neven2018towards}. Recently, various methods such as anchor-based methods~\cite{tabelini2020keep, chen2019pointlanenet, li2020curvelane}, row-wise detection methods~\cite{qin2020ultra, philion2019fastdraw, yoo2020end}, and parametric prediction methods~\cite{tabelini2020polylanenet, liu2021end} have been proposed and continue to refresh the accuracy and efficiency.

\begin{figure}[t!]
\centering
\includegraphics[width=1\linewidth]{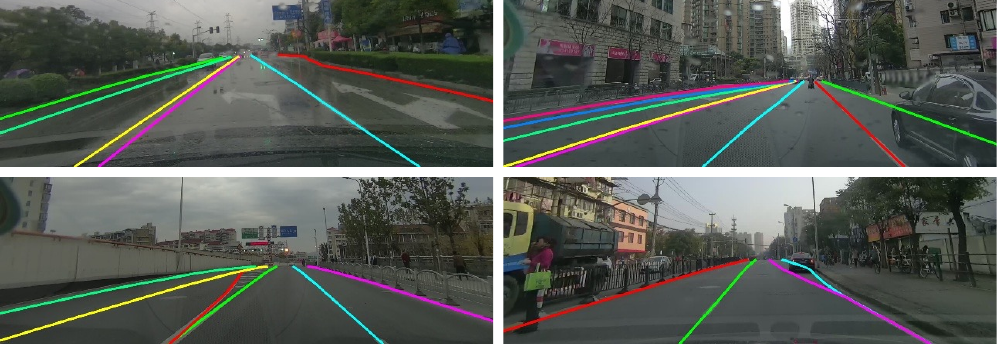}
\caption{Scenes of lane lines with complex topologies. It is challenging to cope with the scenes such as the dense lines(the first row) and the fork lines(the second row). Different instances are represented by different colors in this figure.} %最终文档中希望显示的图片标题
\label{Fig.first} %用于文内引用的标签
\end{figure}
Although deep-learning-based lane detection methods have made great progress~\cite{yurtsever2020survey}, there are still many challenges.

A common problem for lane detection is instance-level discrimination. Most lane detection methods~\cite{neven2018towards, pan2018spatial, ko2020key, tabelini2020keep, hou2019learning, lee2017vpgnet, philion2019fastdraw, chen2019pointlanenet, qin2020ultra, yoo2020end, li2020curvelane} predict lane points first and then aggregate the points into lines. But it is still a common challenge to assign different points to different lane instances~\cite{tang2020review}. A simple solution is to label the lane lines into classes of a fixed number(e.g. labeled as 0, 1, 2, 3 if the maximum lane number is 4) and make a multi-class classification~\cite{pan2018spatial, qin2020ultra, yoo2020end, chougule2018reliable}. But the limitation is that only a predefined, fixed number of lanes can be detected~\cite{neven2018towards}. To overcome this limitation, the post-clustering strategy is investigated~\cite{neven2018towards, ko2020key}. However, this strategy is struggling for some cases such as dense lines. Another approach is anchor-based methods~\cite{liu2021end, li2019line, li2020curvelane}. But it is not flexible to predict the line shape due to the fixed shape of the anchor~\cite{li2020curvelane}.

Another challenge is the detection of lane lines with complex topologies, such as fork lines and dense lines, as is shown in Figure ~\ref{Fig.first}. Such cases are common in driving scenarios, e.g. fork lines usually appear when the number of lanes changes. Homayounfar et al.\cite{homayounfar2019dagmapper} proposed an offline lane detection method for HDMap(High Definition Map) which can deal with the fork lines. However, there are few studies on the perception of lane lines with complex topologies for real-time driving scenarios.

The lane detection task is similar to instance segmentation, which requires assigning different pixels to different instances. Recently, some studies~\cite{tian2020conditional, wang2020solov2} have investigated the conditional instance segmentation strategy, which is also promising for lane detection tasks. However, it is inefficient to directly apply this strategy to lane detection, since the constraint for the mask is not completely consistent with specifying the line shape.~\cite{chougule2018reliable, tang2020review, qin2020ultra}. 

In this work, we propose a novel lane detection framework called CondLaneNet. Aiming to resolve the lane instance-level discrimination problem, we propose the conditional lane detection strategy inspired by CondInst~\cite{tian2020conditional} and SOLOv2~\cite{wang2020solov2}. Different from the instance segmentation tasks, we focus the optimization on specifying the lane line shape based on the row-wise formulation~\cite{qin2020ultra, yoo2020end}. Moreover, we design the Recurrent Instance Module(RIM) to deal with the detection of lane lines with complex topologies such as the dense lines and fork lines. Besides, benefit from the end-to-end pipeline that requires little post-process, our method achieves real-time efficiency. The contributions of this work are summarised as follows:

\begin{itemize}
    \item We have greatly improved the ability of lane instance-level discrimination by the proposed conditional lane detection strategy and row-wise formulation.
    \item We solve the problem of detecting lane lines with complex topologies such as dense lines and fork lines by the proposed RIM. 
    \item Our CondLaneNet framework achieves state-of-the-art performance on multiple datasets, e.g. an 86.10 F1 score(4.6\% higher than SOTA) on CurveLanes and a 79.48 F1 score(3.2\% higher than SOTA) on CULane. Moreover, the small version of our CondLaneNet has high efficiency while ensuring high accuracy, e.g. a 78.14 F1 score and 220 FPS on CULane.
\end{itemize}

%------------------------------------------------------------------------

\section{Related Work}
This section introduces the recent deep-learning-based lane detection methods. According to the strategy of line shape description, current methods can be divided into four categories: segmentation-based methods, anchor-based methods, row-wise detection methods, and parametric prediction methods.

\subsection{Segmentation-based Methods}

Segmentation-based methods~\cite{pan2018spatial,hou2019learning, neven2018towards, ko2020key, lee2017vpgnet, ghafoorian2018gan} are most common and have achieved impressive performance. 
Different from general semantic segmentation tasks, lane detection requires instance-level discrimination. Early methods~\cite{pan2018spatial,ghafoorian2018gan} used a multi-class classification strategy for lane instance discrimination. As explained in the previous section, this strategy is inflexible. 
For higher instance accuracy, the post-clustering strategy~\cite{de2017semantic} was widely applied~\cite{neven2018towards,ko2020key}.
Considering that the segmentation-based methods generally predict a down-scaled mask, some methods~\cite{ko2020key} predict an offset map for refinement.
Recently, some studies~\cite{chougule2018reliable, qin2020ultra} indicated that it is inefficient to describe the lane line as a mask because the emphasis of segmentation is obtaining accurate classification per pixel rather than specifying the line shape. 
To overcome this problem, anchor-based methods and row-wise detection methods were proposed.

\subsection{Anchor-based Methods}
Anchor-based methods~\cite{tabelini2020keep, chen2019pointlanenet, li2020curvelane} take a top-to-down pipeline and focus the optimization on the line shape by regressing the relative coordinates.  
The predefined anchors can reduce the impact of the no-visual-clue problem~\cite{tabelini2020keep} and improve the ability of instance discrimination. 
Due to the slender shape of lane lines, the widely used box-anchor in object detection~\cite{he2017mask} cannot be used directly.  PointLaneNet~\cite{chen2019pointlanenet} and CurveLane~\cite{li2020curvelane} used vertical lines as anchors. LaneATT~\cite{tabelini2020keep} designed anchors with a slender shape and achieves state-of-the-art performance on multiple datasets. However, the fixed anchor shape results in a low degree of freedom in describing the line shape~\cite{li2020curvelane}.

\subsection{Row-wise Detection Methods}
Row-wise detection methods~\cite{qin2020ultra, philion2019fastdraw, yoo2020end} make good use of the shape prior and predict the line location for each row. In the training phase, the constraint on the overall line shape is realized through the location constraint of each row. Based on the continuity and consistency of the predicted locations from row to row, shape constraints can be added to the model~\cite{philion2019fastdraw, qin2020ultra}. Besides, in terms of efficiency, some recent row-wise detection methods\cite{qin2020ultra, yoo2020end, hou2020inter} have achieved advantages.
However, instance-level discrimination is still the main problem for row-wise formulation. 
As the widely used post-clustering
module~\cite{de2017semantic} in segmentation-based methods~\cite{neven2018towards,ko2020key} cannot be directly integrated into the row-wise formulation, row-wise detection methods still take the multi-class classification strategy for lane instance discrimination.
Considering the impressive performance on accuracy and efficiency, we also adopt the row-wise formulation and propose some novel strategies to overcome the instance-level discrimination problem.

\subsection{Parametric Prediction Methods}
Different from the above methods which predict points, parametric prediction methods directly output parametric lines expressed by curve equation. PolyLaneNet~\cite{tabelini2020polylanenet} firstly proposed to use a deep network to regress the lane curve equation. LSTR~\cite{liu2021end} introduced transformer~\cite{vaswani2017attention} to lane detection task and get 420fps detection speed. However, the parametric prediction methods have not surpassed other methods in terms of accuracy.

{\begin{figure*}[ht]
\centering
\includegraphics[width=1\linewidth]{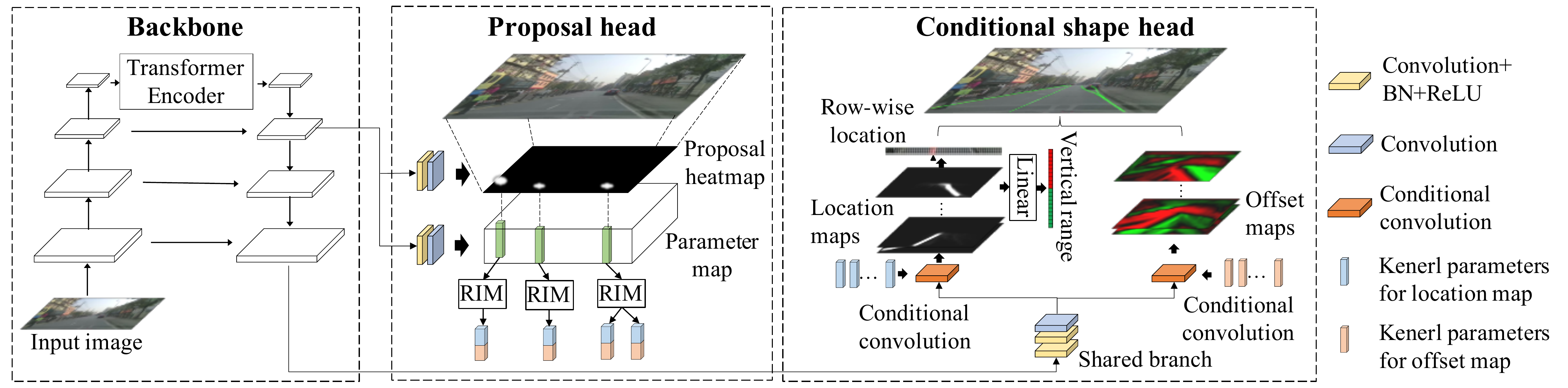}
\caption{The structure of our CondLaneNet framework. The backbone adopts standard ResNet~\cite{he2016deep} and FPN~\cite{lin2017feature} for multi-scale feature extraction. The transformer encoder module~\cite{vaswani2017attention} is added for more efficient context feature extraction. The proposal head is responsible for detecting the proposal points which are located at the start point of the line. Meanwhile, a parameter map that contains the dynamic convolution kernels is predicted. The conditional shape head predicts the row-wise location, the vertical range, and the offset map to describe the shape for each line. To address the cases of dense lines and fork lines, the RIM is designed. }
\label{Fig.main2} %用于文内引用的标签
\end{figure*}}

%%%%%%%%% Introduction
\section{Methods}

Given an input image \(I\in R^{C\times H\times W}\), the goal of our CondLaneNet is to predict a collection of lanes \(L=\left \{ l_1, l_2, ..., l_N \right \}\), where \(N\) is the total number of lanes. Generally, each lane \(l_k\) is represented by an ordered set of coordinates as follows.
\begin{equation}
    \displaystyle
    l_k=[(x_{k1}, y_{k1}), (x_{k2}, y_{k2}), ..., (x_{kN_k}, y_{kN_k})]
\end{equation}%
Where \(k\) is the index of lane and \(N_k\) is the max number of sample points of the \(kth\) lane. 

The overall structure of our CondLaneNet is shown in Figure ~\ref{Fig.main2}.
This section will first present the conditional lane detection strategy, then introduce the RIM(Recurrent Instance Module), and finally detail the framework design.

\subsection{Conditional Lane Detection}

\begin{figure}[ht]
\centering
\includegraphics[width=1\linewidth]{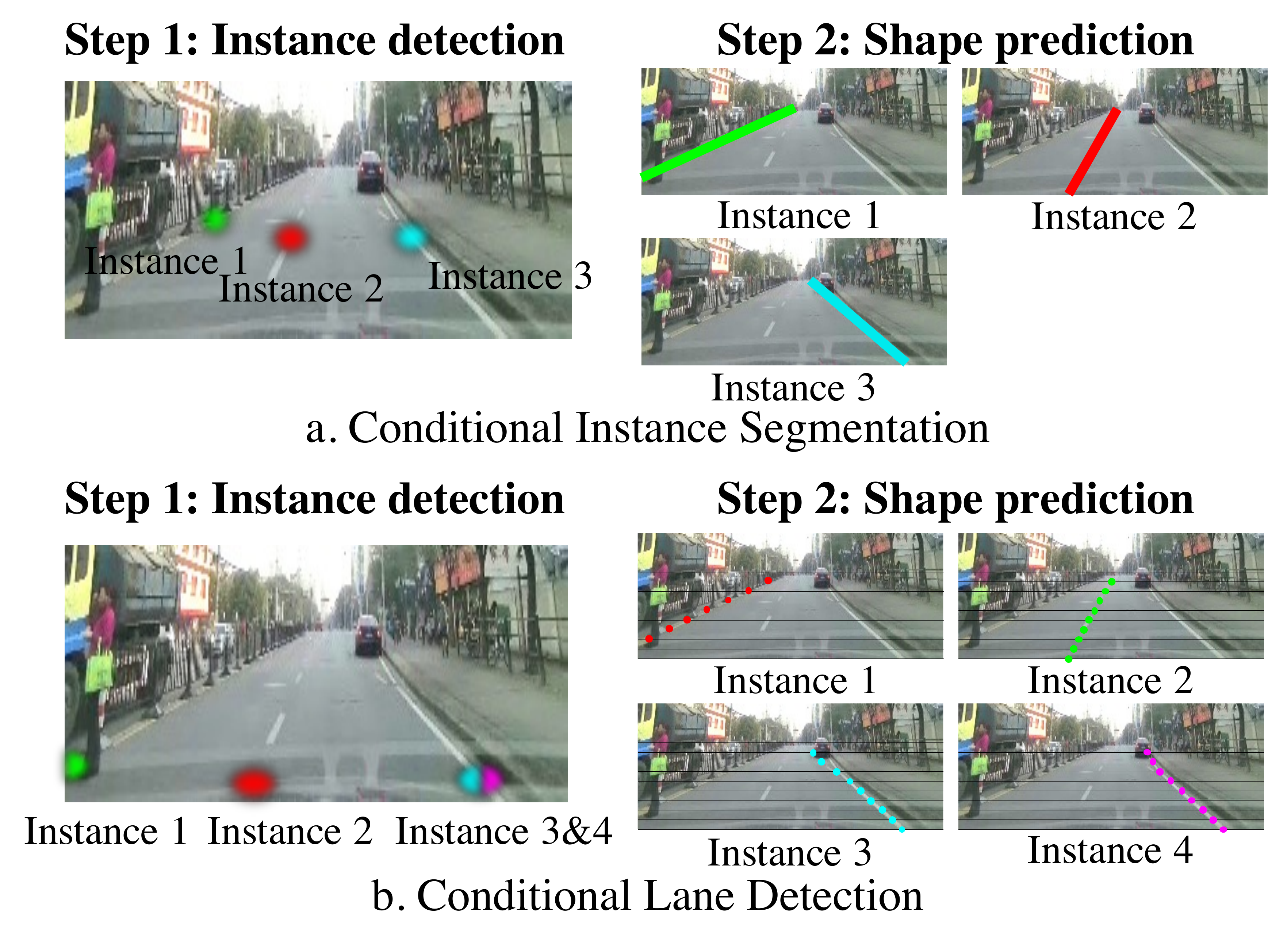}
\caption{The difference between conditional instance segmentation and the proposed conditional lane detection strategy. Our CondLaneNet detects the start point of the lane lines to detect the instance and uses the row-wise formulation to describe the line shape instead of the mask. The overlapping lines can be distinguished based on the proposed RIM, which will be detailed in Section 3.2.} %最终文档中希望显示的图片标题
\label{Fig.compare} %用于文内引用的标签
\end{figure}

Focusing on the instance-level discrimination ability, we propose the conditional lane detection strategy based on conditional convolution -- a convolution operation with dynamic kernel parameters~\cite{jia2016dynamic,yang2019condconv}. The conditional detection process~\cite{tian2020conditional, wang2020solov2} has two steps: instance detection and shape prediction, as is shown in Figure ~\ref{Fig.compare}. The instance detection step predicts the object instance and regresses a set of dynamic kernel parameters for each instance. In the shape prediction step, conditional convolutions are applied to specify the instance shape. This process is conditioned on the dynamic kernel parameters. Since each instance corresponds to a set of dynamic kernel parameters, the shapes can be predicted instance-wisely.

This strategy has achieved impressive performance on instance segmentation tasks~\cite{tian2020conditional, wang2020solov2}. However, directly applying the conditional instance segmentation strategy to lane detection is blunt and inappropriate. On the one hand, the segmentation-based shape description is inefficient for lane lines due to the excessively high degree of freedom~\cite{qin2020ultra}. On the other hand, the instance detection strategy for general objects is not suitable for slender and curved objects due to the inconspicuous visual characteristic of the border and the central. Our conditional lane detection strategy improves shape prediction and instance detection to address the above problems.

\subsubsection{Shape Prediction}
We improve the row-wise formulation~\cite{qin2020ultra} to predict the line shape based on our conditional shape head, as is shown in Figure ~\ref{Fig.main2}. In the row-wise formulation, we predict the lane location on each row and then 
aggregate the locations to get the lane line in the order from bottom to top, based on the prior of the line shape. Our row-wise formulation has three components: the row-wise location, the vertical range, and the offset map. The first two outputs are basic elements for most row-wise detection methods~\cite{qin2020ultra, yoo2020end}. Besides, we predict an offset map as the third output for further refinement.

\begin{figure}[ht]
\centering
\includegraphics[width=1\linewidth]{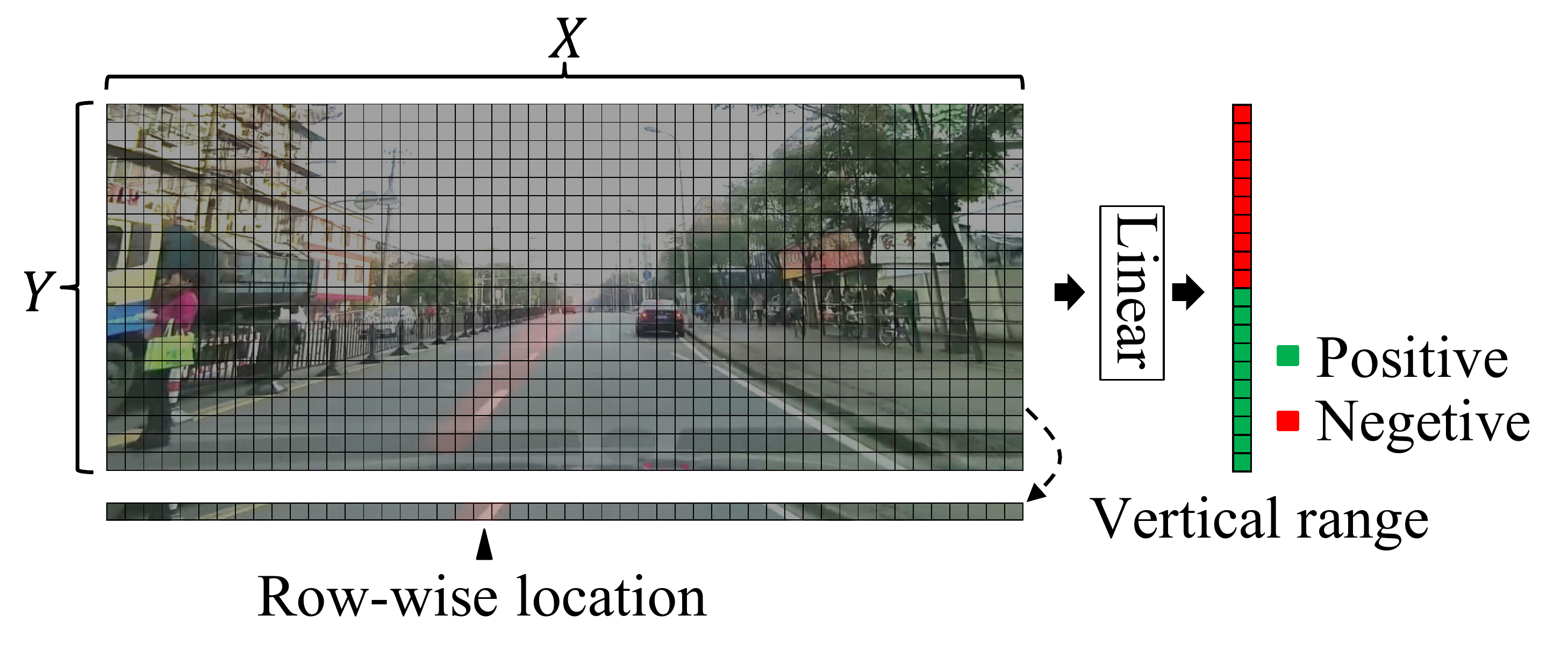}
\caption{The process of parsing the row-wise location and the vertical range from the location map.} %最终文档中希望显示的图片标题
\label{Fig.row} %用于文内引用的标签
\end{figure}

\paragraph{Row-wise Location}
As is shown in Figure ~\ref{Fig.row}, we divide the input image into grids of shape \(Y\times X\) and predict a corresponded location map, which is a feature map of shape \(1\times Y\times X\) output by the proposed conditional shape head. On the location map, each row has an abscissa indicating the location of the lane line.

To get the row-wise location, a basic approach is to process the \(X\)-classes classification in each row. In inference time, the row-wise location is determined by picking the most responsive abscissa in each row. However, a common situation is that the line location is between the two grids, and both the two grids should have a high response. To overcome this problem, we introduce the following formulation.

For each row, we predict the probability that the lane line appears in each grid.

\begin{equation}
    \displaystyle
    {p_i=softmax(f_{loc}^i)}
\end{equation}%
Where \(i\) represents the \(ith\) row, \(f_{loc}^i\) is the feature vector of the \(ith\) row of location map \(f_{loc}\), \(p_i\) is the probability vector for the \(ith\) row.

The final row-wise location is defined as the expected abscissa.
\begin{equation}
    \displaystyle
{E(\hat x_i)=\sum_{j}^{} j\cdot p_{ij}}
\label{Eq.location} %用于文内引用的标签
\end{equation}%

 Where \(E(\hat x_i)\) is the expected abscissa, \(p_{ij}\) is the probability of the lane line passing through the coordinate \((j, i)\).

In the training phase, L1-loss is applied.
\begin{equation}
    \displaystyle
\ell _{row} = \frac{1}{N_v}\sum_{i\in V}^{}  |E(\hat x_i)- x_i|
\end{equation}%
Where \(V\) represents the vertical range of the labeled line, \(N_v\) is the number of valid rows.

\paragraph{Vertical Range}
The vertical lane range is determined by row-wisely predicting whether the lane line passes through the current row, as is shown in Figure ~\ref{Fig.row}. We add a linear layer and perform binary-classification row by row. We use the feature vector of each row in the location map as the input. The softmax-cross-entropy loss is adopted to guide the training process.

\begin{equation}
\displaystyle
\ell _{range}=\sum_{i}^{}( -y_{gt}^ilog(v_i)-(1-y_{gt}^i)log(1-v_i))
\end{equation}%
Where \(v_i\) represents the predicted positive probability for \(ith\) row and \(y_{gt}^i\) is the groundtruth of \(ith\) row.

\paragraph{Offset Map}
The row-wise location defined in Equation~\ref{Eq.location} points to the abscissa of the vertex on the left side of the grid, rather than the precise location. Thus, we add the offset map to predict the offset in the horizontal direction near the row-wise location for each row. We use L1-loss to constrain the offset map as follows.
\begin{equation}
    \displaystyle
    \ell_{offset}=\frac{1}{N_{\Omega}} \sum_{(j,i)\in \Omega }^{} \left | \hat{\delta}_{ij} - \delta_{ij}  \right | 
\end{equation}%
where \(\hat{\delta}_{ij}\) and \(\delta_{ij}\) are the predicted offset and the label offset on coordinate \((j,i)\). We define \(\Omega\) as the area near the lane line with a fixed width. \(N_{\Omega}\) is the number of pixels in \(\Omega\).

\paragraph{Shape Description}
Each output lane line is represented as an ordered set of coordinates.
For \(kth\) line, the coordinate \((x_k^i, y_k^i)\) of the \(ith\) row is represented as follows.
\begin{equation}
    \displaystyle
    \left\{\begin{array}{l}
    y_k^i=H/Y\cdot i 
    \\
    x_k^i=W/X \cdot (loc_k^i+\delta (loc_k^i, i))
    
\end{array}\right.
\end{equation}%
Where \(i\in \left [ v_{min}^{k}, v_{max}^{k} \right ]\), \(v_{min}^{k}\) and \(v_{max}^{k}\) are respectively the minimum and maximum values of the predicted vertical range,
\(loc_{i}^{k}\) is rounded down from \(E_{i}^{k}\), \(\delta(\cdot)\) is the predicted offset. 

\subsubsection{Instance Detection}
We design the proposal head for instance detection, as is shown in Figure ~\ref{Fig.main2}.
For general conditional instance segmentation methods~\cite{tian2020conditional, wang2020solov2}, the instance is detected in an end-to-end pipeline by predicting the central of each object. 
However, it is hard to predict the central for the slender and curved lines because the visual characteristic of the line central is not obvious.

We detect the lane instance by detecting the proposal point located at the start point of the line. The start point has a more clear definition and more obvious visual characteristic than the central. We follow CenterNet~\cite{duan2019centernet} and predict a proposal heatmap to detect the proposal points.
To constraint the proposal heatmap, we adopt focal loss following CornerNet~\cite{law2018cornernet} and CenterNet\cite{duan2019centernet}.
\begin{equation}
\resizebox{.91\linewidth}{!}{$
    \displaystyle
    \ell_{point}=\frac{-1}{N_p}\sum_{xy}^{}\begin{cases} (1-\hat{P}_{xy})^\alpha  log(\hat{P}_{xy})&{P_{xy}=1} \\
    (1-P_{xy})^\beta({\hat{P}_{xy}})^{\alpha}log(1-\hat{P}_{xy})   & otherwise
\end{cases}
$}
\end{equation}

Where \(P_{xy}\) is the label at coordinate \((x,y)\) and \(\hat{P}_{xy}\) is the predicted value at coordinate \((x,y)\) of the proposal heatmap. \(N_p\) is the number of proposal points in the input image.

Besides, we regress the dynamic kernel parameters by predicting a parameter map following CondInst~\cite{tian2020conditional} and SOLOv2~\cite{wang2020solov2}. The constraints of the parameter map are constructed through the constraints on the line shape.

\subsection{Recurrent Instance Module}
In the proposal head described above, each proposal point is bound to a lane instance. However, in practice, multiple lane lines can fall in the same proposal point such as the fork lanes. To deal with the above cases, we propose the Recurrent Instance Module(RIM).

\begin{figure}[h]
\begin{center}
\includegraphics[scale=0.4]{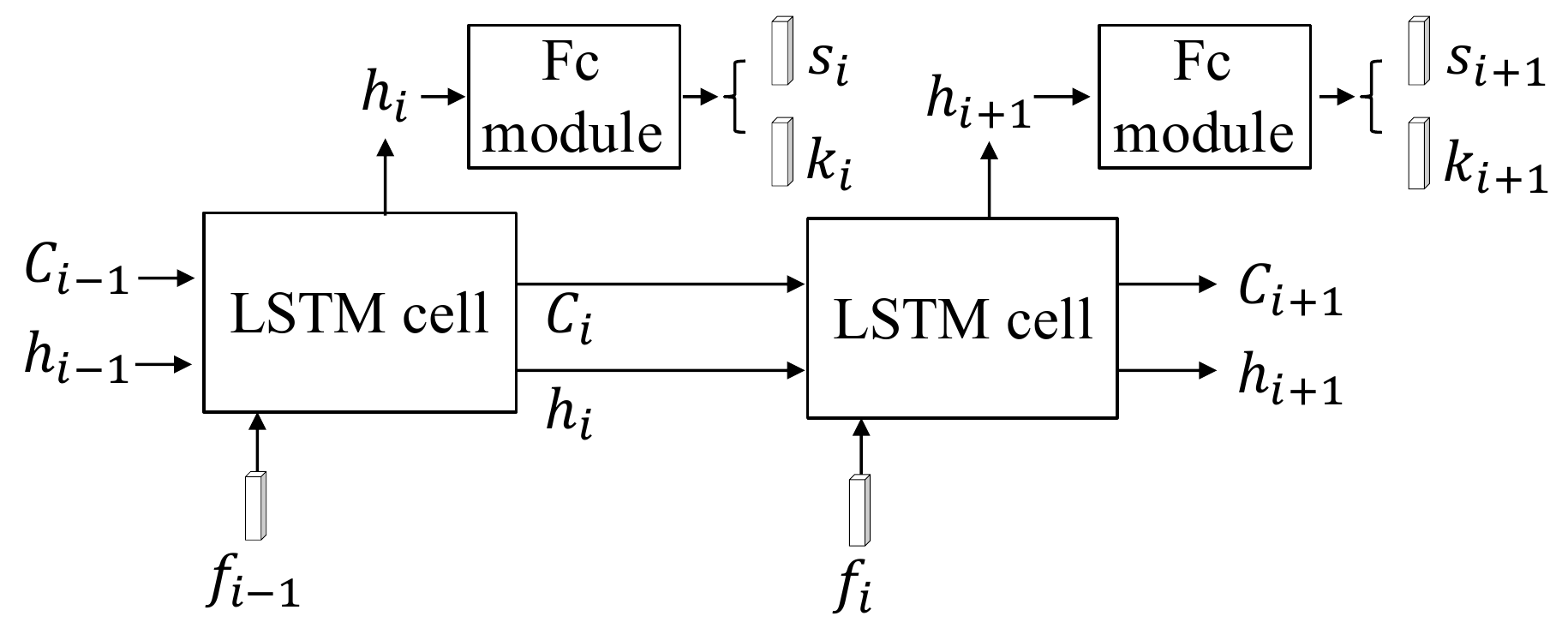}
\end{center}
   \caption{The Recurrent Instance Module. In this figure, \(h\) and \(c\) are the short-term memory and long-term memory respectively, \(f\) is the input feature vector, \(s\) is the output state logit, \(k\) is the output kernel parameter vector.}
\label{Fig.lstm}
\end{figure}
The structure of the proposed RIM is shown in Figure ~\ref{Fig.lstm}. Based on LSTM(Long Short-term Memory)~\cite{hochreiter1997long}, the RIM recurrently predicts a state vector \(s_i\) and a kernel parameter vector \(k_i\). We define \(s_i\)  as two-dimensional logits that indicate two states: ``continue'' or ``stop''. The vector \(k_i\) contains the kernel parameters for subsequent instance-wise dynamic convolution. In the inference phase, the RIM recurrently predicts the lane-wise kernel parameters bound to the same proposal point until the state is ``stop''. As is shown in Figure \ref{Fig.main2}, RIM is added for each proposal point. Therefore, each proposal point can guide the shape prediction of multiple lane instances.

We adopt cross-entropy loss to constrain the state output as follows.

\begin{equation}
    \displaystyle
    \ell_{state}=\frac{1}{N_s}\sum_{i}^{}-\left [ y_i\cdot log(s_i) + (1-y_i)\cdot log(1-s_i)\right ] 
\end{equation}%

Where \(s_i\) is the output of softmax operation for \(ith\) state, result \(y_i\) is the ground truth for the \(ith\) state and \(N_s\) is the total number of the state outputs in a batch.

In the training phase, the total loss is defined as follows.
\begin{equation}
    \displaystyle
\ell _{total} = \ell _{point}+\alpha \ell _{row}+\beta \ell _{range}+\gamma \ell _{offset}+\eta \ell _{state}
\end{equation}%

The hyperparameters \(\alpha\), \(\beta\), \(\gamma\) and \(\eta\) are set to 1.0, 1.0, 0.4 and 1.0 respectively.

\subsection{Architecture}
The overall architecture is shown in Figure ~\ref{Fig.main2}. We adopt ResNet~\cite{he2016deep} as the backbone and add a standard FPN~\cite{lin2017feature} module to provide integrated multi-scale features. The proposal head detects the lane instances by predicting the proposal heatmap of shape \(1 \times H_p \times W_p\). Meanwhile, a parameter map of shape \(C_p \times H_p \times W_p\) that contains the dynamic kernel parameters is predicted. For the instance with the proposal point located at \((x_p, y_p)\), the corresponding dynamic kernel parameters are contained in the \(C_p\) dimensional kernel feature vector at \((x_p, y_p)\) on the parameter map. Further, given the kernel feature vector, the RIM recurrently predicts the dynamic kernel parameters.
Finally, the conditional shape head predicts the line shape instance-wisely conditioned on the dynamic kernel parameters.

\begin{figure}[h]
\centering
\includegraphics[scale=0.45]{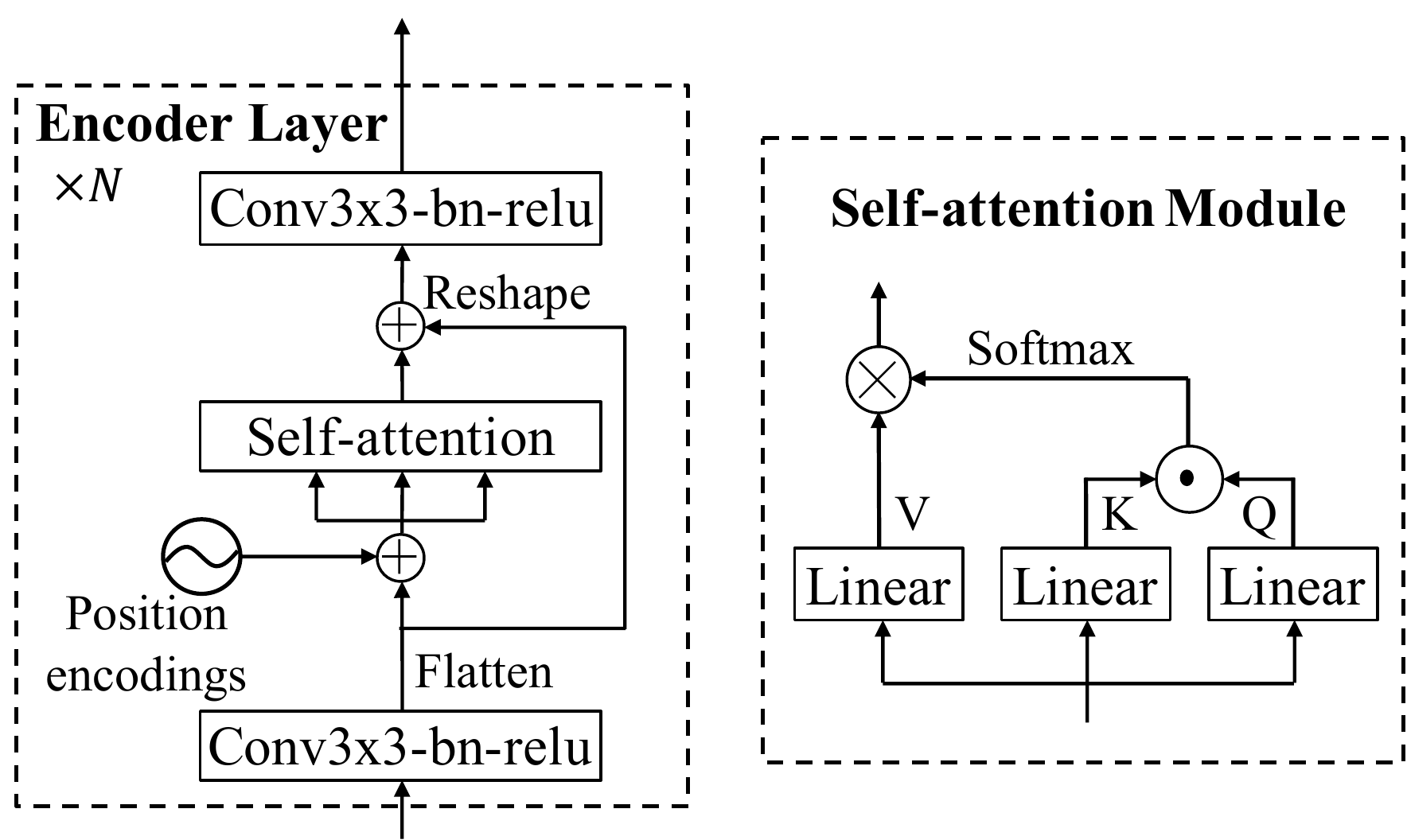}
\caption{The structure of the transformer encoder. The \(\oplus\), \(\odot\) and \(\otimes\) respectively represent matrix addition, dot-product operation and element-wise product operation.} %最终文档中希望显示的图片标题
\label{Fig.encoder} %用于文内引用的标签
\end{figure}
Our framework requires a strong capability of context feature fusion. For example, the prediction of the proposal point is based on the features of the entire lane line which generally has an elongated shape and long-range. Therefore, we add a transformer encoder structure to the last layer of the backbone for the fusion of contextual information. We retain the two-dimensional spatial features in the encoder layer and use convolutions for feature extraction. The structure of the transformer encoder used in our framework is shown in Figure \ref{Fig.encoder}.

%%%%%%%%% Experiments
\section{Experiments}
\subsection{Experimental Setting}
\subsubsection{Datasets}
To extensively evaluate the proposed method, we conducte experiments on three benchmarks: CurveLanes~\cite{li2020curvelane}, CULane~\cite{pan2018spatial}, and TuSimple~\cite{c.elmohamed}. CurveLanes is a recently proposed benchmark with cases of complex topologies such as fork lines and dense lines. CULane is a widely used large lane detection dataset with 9 different scenarios. TuSimple is another widely used dataset of highway driving scenes. The details of the three datasets are shown in Tab. ~\ref{tab:1}.
\begin{table}[h]
\begin{center}
\scalebox{0.82}{
\begin{tabular}{lccccc}
\hline
Dataset    & Train & Val. & Test &Road type & Fork \\
\hline
CurveLanes & 100K & 20K & 30K &Urban$\&$Highway  & $\surd$ \\
CULane & 88.9K & 9.7K & 34.7K &Urban$\&$Highway  & $\times$ \\
TuSimple & 3.3K & 0.4K & 2.8K &Highway  & $\times$ \\
\hline
\end{tabular}}
\end{center}
\caption{Details of three datasets.}
\label{tab:1}
\end{table}

\subsubsection{Evaluation Metrics}
For CurveLanes and CULane, we adopte the evaluation metrics of SCNN~\cite{pan2018spatial} which utilizes the F1 measure as the metric. IoU between the predicted lane line and GT label is taken for judging whether a sample is true positive (TP) or false positive (FP) or false negative (FN). IoU of two lines is defined as the IoU of their masks with a fixed line width. Further, F1-measure is calculated as follows:
\begin{equation}
    \displaystyle
Precision=\frac{TP}{TP+FP}
\label{pr} 
\end{equation}%

\begin{equation}
    \displaystyle
Recall=\frac{TP}{TP+FN}
\label{re} 
\end{equation}%

\begin{equation}
    \displaystyle
F1=\frac{2\times Precision\times Recall}{Precision+Recall}
\label{f1} 
\end{equation}%

For TuSimple dataset~\cite{c.elmohamed}, there are three official indicators: false-positive rate (FPR), false-negative rate (FNR), and accuracy.
\begin{equation}
    \displaystyle
accuracy = \frac{\sum_{clip} C_{clip}}{\sum_{clip} S_{clip}}
\end{equation}%
Where \(C_{clip}\) is the number of correctly predicted lane points and \(S_{clip}\) is the total number of lane points of a clip. 
Lane with accuracy greater than 85$\%$ is considered as a true-positive otherwise false positive or false negative. Besides, the F1 score is also reported.

\subsubsection{Implementation details}

We fix the large, medium, and small versions of our CondLaneNet for all three datasets. The difference between the three models is shown in Table ~\ref{tab:difference}.
For all three datasets, input images are resized to 800\(\times\)320 pixels during training and testing. Since there are no cases of fork lines in CULane and TuSimple, RIM is only applied for the CurveLanes dataset. In the optimizing process, we use Adam optimizer~\cite{kingma2014adam} and step learning rate decay~\cite{loshchilov2017decoupled} with an initial learning rate of 3e-4. For each dataset, we train on the training set without any extra data. We respectively train 14, 16 and 70 epochs for CurveLanes, CULane and TuSimple with a batchsize of 32.
The results are reported on the test set for CULane and TuSimple. For CurveLanes, we report the results on the validation set following CurveLane~\cite{li2020curvelane}. All the experiments were computed on a machine with an RTX2080 GPU.
\begin{table}[!h]
\centering
\scalebox{0.8}{
\begin{tabular}{lccc}
\hline
Model name    & Backbone & \makecell[c]{Proposal head input} & \makecell[c]{Shape head input} \\
\hline
Large & Resnet-101 & downscale 16 & downscale 4 \\
Medium & Resnet-34 & downscale 16 & downscale 8 \\
Small & Resnet-18 & downscale 16 & downscale 8 \\
\hline
\end{tabular}}
\caption{Difference of different versions of our CondLaneNet.}
\label{tab:difference}
\end{table}

{\begin{figure*}[ht]
\centering
\includegraphics[scale=0.48]{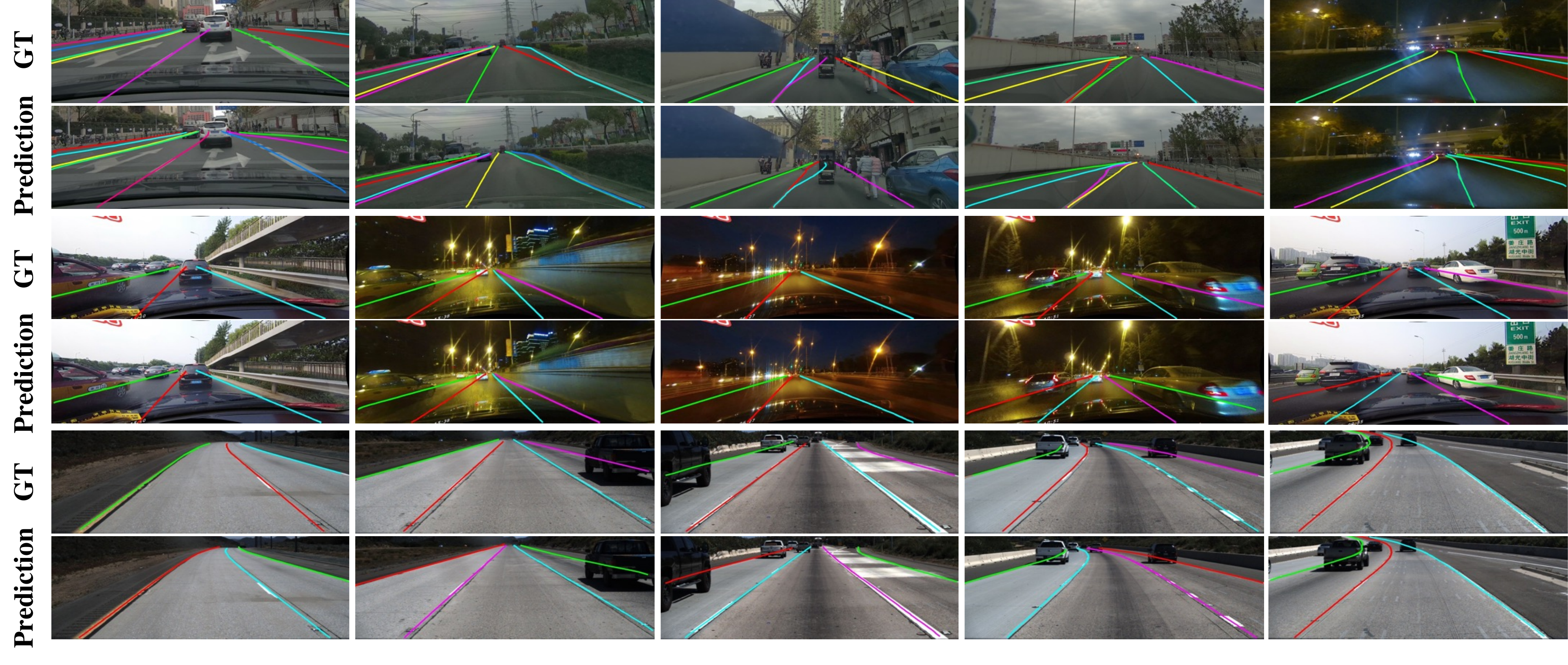}
\caption{Visualization results on CurveLanes(the first row), CULane(the middle row) and TuSimple(the last row) datasets. Different lane instances are represented by different colors.
}
\label{Fig.vis} %用于文内引用的标签
\end{figure*}}
\subsection{Results}
The visualization results on the CurveLanes, CULane, and TuSimple datasets are shown in the Figure ~\ref{Fig.vis}.
The results show that our method can cope with complex line topologies. Even for the cases of dense lines and fork lines, our method can also successfully discriminate the instances.

\paragraph{CurveLanes}
\begin{table}[h]
\centering
\scalebox{0.8}{
\begin{tabular}{lrrrrr}
\hline
Method    & F1 & Precision & Recall & FPS & GFlops(G) \\
\hline
SCNN~\cite{pan2018spatial} & 65.02 & 76.13 & 56.74 &  & 328.4 \\
Enet-SAD~\cite{hou2019learning} & 50.31 & 63.60 & 41.60 &  & \textbf{3.9} \\
PointLaneNet~\cite{chen2019pointlanenet} & 78.47 & 86.33 & 72.91 &  & 14.8 \\
CurveLane-S~\cite{li2020curvelane} & 81.12 & \textbf{93.58} & 71.59 &  & 7.4 \\
CurveLane-M~\cite{li2020curvelane}  & 81.80 & 93.49 & 72.71 &  & 11.6 \\
CurveLane-L~\cite{li2020curvelane}  & 82.29 & 91.11 & 75.03 &  & 20.7 \\
\hline
CondLaneNet-S & 85.09 & 87.75 & 82.58 & 154 & 10.3 \\
CondLaneNet-M & 85.92 & 88.29  & \textbf{83.68}  & 109 & 19.7    \\
CondLaneNet-L & \textbf{86.10} & 88.98 & 83.41 & 48 & 44.9  \\
\hline
\end{tabular}}
\caption{Comparison of different methods on CurveLanes.}
\label{tab:curvelanes}
\end{table}

The comparison results on CurveLanes are shown in Tabel ~\ref{tab:curvelanes}. CurveLanes contains cases of lane lines with complex topologies such as curve, fork, and dense lanes. Our large version of CondLaneNet achieves a new state-of-the-art F1 score of 86.10, 4.63\% higher than CurveLane-L. Our small version of CondLaneNet still has a performance of an 85.09 F1 score (3.40\% higher than SOTA). Since our model can deal with cases of the fork and dense lane lines, there is a significant improvement in the recall indicator. Correspondingly, false-positive results will increase, resulting in a decrease in the precision indicator.

\paragraph{CULane}
\begin{table*}[!b]
\centering
\scalebox{0.82}{
\begin{tabular}{lrrrrrrrrrrrr}
\hline
Category    & Total & Normal & Crowded & Dazzle & Shadow & No line & Arrow & Curve & Cross & Night & FPS & GFlops(G) \\
\hline
SCNN~\cite{pan2018spatial} & 71.60 & 90.60 & 69.70 & 58.50 & 66.90 & 43.40 & 84.10 & 64.40 & 1990 & 66.10 & 7.5 & 328.4     \\
ERFNet-E2E~\cite{yoo2020end} & 74.00 & 91.00 & 73.10 & 64.50 & 74.10 & 46.60 & 85.80 & 71.90 & 2022 & 67.90 & &    \\
FastDraw~\cite{philion2019fastdraw} & & 85.90 & 63.60 & 57.00 & 69.90 & 40.60 & 79.40 & 65.20 & 7013 & 57.80 & 90.3 &   \\
ENet-SAD~\cite{hou2019learning} & 70.80 & 90.10 & 68.80 & 60.20 & 65.90 & 41.60 & 84.00 & 65.70 & 1998 & 66.00 & 75 & \textbf{3.9}      \\
UFAST-ResNet34~\cite{qin2020ultra} & 72.30 & 90.70 & 70.20 & 59.50 & 69.30 & 44.40 & 85.70 & 69.50 & 2037 & 66.70 & 175.0 & \\  
UFAST-ResNet18~\cite{qin2020ultra} & 68.40 & 87.70 & 66.00 & 58.40 & 62.80 & 40.20 & 81.00 & 57.90 & 1743 & 62.10 &  \textbf{322.5} & \\  
ERFNet-IntRA-KD~\cite{hou2020inter} & 72.40 &  &  &  &  &  &  &  &  &  &100.0  & \\ 
CurveLanes-NAS-S~\cite{li2020curvelane} & 71.40 & 88.30 & 68.60 & 63.20 & 68.00 & 47.90 & 82.50 & 66.00 & 2817 & 66.20 &  &  9.0      \\
CurveLanes-NAS-M~\cite{li2020curvelane} & 73.50 & 90.20 & 70.50 & 65.90 & 69.30 & 48.80 & 85.70 & 67.50 & 2359 & 68.20 &  & 35.7      \\
CurveLanes-NAS-L~\cite{li2020curvelane} & 74.80 & 90.70 & 72.30 & 67.70 & 70.10 & 49.40 & 85.80 & 68.40 & 1746 & 68.90 &  & 86.5      \\
LaneATT-Small~\cite{tabelini2020keep} & 75.13 & 91.17 & 72.71 & 65.82 & 68.03 & 49.13 & 87.82 & 63.75 & \textbf{1020} & 68.58 & 250 & 9.3     \\
LaneATT-Medium~\cite{tabelini2020keep} & 76.68 & 92.14 & 75.03 & 66.47 & 78.15 & 49.39 & 88.38 & 67.72 & 1330 & 70.72 & 171 & 18.0      \\
LaneATT-Large~\cite{tabelini2020keep} & 77.02 & 91.74 & 76.16 & 69.47 & 76.31 & 50.46 & 86.29 & 64.05 &  1264 & 70.81 & 26 & 70.5      \\
\hline
CondLaneNet-Small & 78.14 & 92.87& 75.79 & 70.72 & 80.01 & 52.39 & 89.37 & 72.40 & 1364 & 73.23 & 220 & 10.2      \\
CondLaneNet-Medium & 78.74 & 93.38 & 77.14 &  \textbf{71.17} & 79.93 &  51.85 & 89.89 &  73.88 & 1387 & 73.92 & 152 & 19.6     \\
CondLaneNet-Large &  \textbf{79.48} &  \textbf{93.47} &  \textbf{77.44} & 70.93 &  \textbf{80.91} & \textbf{54.13} & \textbf{90.16} & \textbf{75.21} & 1201 &  \textbf{74.80} & 58 & 44.8      \\
\hline
\end{tabular}}
\caption{Comparison of different methods on CULane.}
\label{tab:culane}
\end{table*}
The results of our CondLaneNet and other state-of-the-art methods on CULane are shown in Tabel ~\ref{tab:culane}. Our method achieves a new state-of-the-art result of a 79.48 F1 score, which has increased by 3.19\%. Moreover, our method achieves the best performance in eight of nine scenarios, showing robustness to different scenarios. For some hard cases such as curve and night, our methods have obvious advantages. Besides, the small version of our CondLaneNet gets a 78.14 F1 score with a speed of 220 FPS, 1.12 higher and 8.5\(\times\) speed than LaneATT-L. Compared with LaneATT-S, CondLaneNet-S achieves a 4.01 \% F1 score improvement with similar efficiency. In most scenarios of CULane, the small version of our CondLaneNet exceeds all previous methods in the F1 measure.

\paragraph{Tusimple}
The results on TuSimple are shown in Table ~\ref{tab:tusimple}. Relatively, the gap between different methods on this dataset is smaller, due to the smaller amount of data and more single scenes. Our method achieves a new state-of-the-art F1 score of 97.24. Besides, the small version of our method gets a 97.01 F1 score with 220 FPS.

\begin{table}[h]
\centering
\setlength{\tabcolsep}{2.5pt}{
\scalebox{0.81}{
\begin{tabular}{lrrrrrr}
\hline
Method    & F1 & Acc & FP & FN  & FPS & GFLOPS \\
\hline
SCNN~\cite{pan2018spatial} & 95.97 & 96.53 & 6.17 & \textbf{1.80}  & 7.5 & \\
EL-GAN~\cite{ghafoorian2018gan} & 96.26 & 94.90 & 4.12 & 3.36  & 10.0 & \\
PINet~\cite{ko2020key} & 97.21 & 96.70 & 2.94 & 2.63  & & \\
LineCNN~\cite{li2019line} & 96.79 & \textbf{96.87} & 4.42 & 1.97 & 30.0 & \\
PointLaneNet~\cite{chen2019pointlanenet} & 95.07 & 96.34 & 4.67 & 5.18 & 71.0 & \\
ENet-SAD~\cite{hou2019learning} & 95.92 & 96.64 & 6.02 & 2.05 & 75.0 & \\
ERF-E2E~\cite{yoo2020end} & 96.25 & 96.02 & 3.21 & 4.28 & & \\
FastDraw~\cite{philion2019fastdraw} & 93.92 & 95.20 & 7.60 & 4.50 & 90.3 & \\
UFAST-ResNet34~\cite{qin2020ultra} & 88.02 & 95.86 & 18.91 & 3.75  & 169.5 & \\
UFAST-ResNet18~\cite{qin2020ultra} & 87.87 & 95.82 & 19.05 & 3.92  & 312.5 & \\
PolyLaneNet~\cite{tabelini2020polylanenet} & 90.62 & 93.36 & 9.42 & 9.33  & 115.0 & 0.9 \\
LSTR~\cite{liu2021end} & 96.86  & 96.18 & 2.91 & 3.38  & \textbf{420} & \textbf{0.3} \\
LaneATT-ResNet18~\cite{tabelini2020keep} & 96.71 & 95.57 & 3.56 & 3.01  & 250 & 9.3 \\
LaneATT-ResNet34~\cite{tabelini2020keep} & 96.77 & 95.63 & 3.53 & 2.92  & 171 & 18.0 \\
LaneATT-ResNet122~\cite{tabelini2020keep} & 96.06 & 96.10 & 5.64 & 2.17  & 26 & 70.5 \\
\hline
CondLaneNet-S & 97.01 & 95.48 & 2.18  & 3.80 & 220 & 10.2 \\
CondLaneNet-M & 96.98 & 95.37 & 2.20  & 3.82 & 154 & 19.6 \\
CondLaneNet-L & \textbf{97.24} & 96.54 & \textbf{2.01}  & 3.50 & 58 & 44.8 \\

\hline
\end{tabular}}}
\caption{Comparison of different methods on TuSimple.}
\label{tab:tusimple}
\end{table}

\subsection{Ablation Study of Improvement Strategies}
We performed ablation experiments on the CurveLanes dataset based on the small version of our CondLaneNet. The results are shown in Tabel ~\ref{tab:ablation}. We take the lane detection model based on the original conditional instance segmentation strategy~\cite{tian2020conditional, wang2020solov2} (as is shown in Figure ~\ref{Fig.compare}.a) as the baseline.  The first row shows the results of the baseline. In the second row, the proposed conditional lane detection strategy is applied and the lane mask expression is replaced by the row-wise formulation(as is shown in ~\ref{Fig.compare}.b). In the third row, the offset map for post-refinement is added. In the fourth row, the transformer encoder is added and the offset map is removed. The fifth row presents the result of the model with the row-wise formulation, the offset map, and the transformer encoder. In the last row, RIM is added.

\begin{table}[h]
\centering
\scalebox{0.82}{
\begin{tabular}{cccccr}
\hline
Baseline & Row-wise & Offset & Encoder & RIM  & F1 score \\
\hline
$\surd$ &  &  &  &  & 72.19 \\
  & $\surd$ &  &  &  & 80.09(+7.9) \\
  & $\surd$ & $\surd$ &  & & 81.24(+9.05) \\
  & $\surd$ &  & $\surd$ & & 81.85(+9.66) \\
  & $\surd$ & $\surd$ & $\surd$ & & 83.41(+11.22) \\
  & $\surd$ & $\surd$ & $\surd$ & $\surd$ & 85.09(+12.90) \\
\hline
\end{tabular}}
\caption{Ablation study of the improvement strategies on CurveLanes base on the small version of our CondLaneNet.}
\label{tab:ablation}
\end{table}

Comparing the first two rows, we can see that the proposed conditional lane detection strategy has significantly improved the performance. Comparing the results of the 2nd and the 3rd row, the 4th and the 5th row, we can see the positive effect of the offset map. Moreover, the transformer encoder plays a vital role in our framework, which can be indicated by comparing the 2nd and the 4th row, the 3rd and the 5th row. Besides, RIM designed for the fork lines and dense lines also improves the accuracy.

\subsection{Ablation Study of Transformer Encoder}

\begin{table}[h]
\centering
\setlength{\tabcolsep}{2pt}{
\scalebox{0.9}{
\begin{tabular}{l|cc|cc|cc}
\hline
Model & \multicolumn{2}{c|}{Small} &\multicolumn{2}{c|}{Medium} & \multicolumn{2}{c}{Large} \\
\hline
Target & P. point & Line & P. point & Line & P. point & Line \\
\hline
 Standard & 88.35 & 85.09 & 88.99 & 85.92 & 89.54 & 86.10 \\
 S. w/o encoder& 85.51 & 82.97 & 88.68 & 85.91 & 89.33 & 85.98 \\
 Hacked & 88.05 & 84.39 & 88.90 & 85.93 & 89.37 & 85.99 \\
\hline
\end{tabular}}}
\caption{Ablation study of the transformer encoder module on CurveLanes.}
\label{tab:encoder1}
\end{table}
This section further analyzes the function of the transformer encoder which indicates a vital role in the previous experiments. Our method first detects instances by detecting the proposal points and then predicts the shape for each instance. The accuracy of the proposal points greatly affects the final accuracy of the lane lines.
We design different control groups to compare the accuracy of the proposal points and lane lines on CurveLanes. We define the proposal points which locate in the eight neighborhoods of the groundtruth points as the true-positive samples. Considering the function of RIM, the proposal point corresponding to multiple lines are regarded as multiple different proposal points.
We report the F1 score of the proposal points and lane lines, as is shown in Table ~\ref{tab:encoder1}. 

The first row shows the results of the small, medium and large versions of the standard CondLaneNet. In the second row, the transformer encoder is removed. In the third row, we hack the inference process of the second row by replacing the proposal heatmap with the proposal heatmap output by the standard model(the first row). For the small version, removing the encoder leads to a significant drop for both proposal points and lanes. However, using the proposal heatmap of the standard model, the results on the third row are close to the first row. 

The above results prove that the function of the encoder is mainly to improve the detection of the proposal points, which rely on contextual features and global information.
Besides, the contextual features can be more fully refined in deeper networks. Therefore, for the medium and large versions, 
the improvement of the encoder is far less than the small version.

\section{Conclusion}
In this work, We proposed CondLaneNet, a novel top-to-down lane detection framework that detects the lane instances first and then instance-wisely predict the shapes. Aiming to resolve the instance-level discrimination problem, we proposed the conditional lane detection strategy based on conditional convolution and row-wise formulation. Moreover, we designed RIM to cope with complex lane line topologies such as dense lines and fork lines. Our CondLaneNet framework refreshed the state-of-the-art performance on CULane, CurveLanes, and TuSimple. Moreover, on CULane and CurveLanes, the small version of our CondLaneNet not only surpassed other methods in accuracy, but also presented real-time efficiency.

{\small
\bibliographystyle{ieee_fullname}
\bibliography{egbib}
}

\end{document}

%%%%%%%%% TITLE
\title{CondLaneNet: a Top-to-down Lane Detection Framework Based on Conditional Convolution}

\author{First Author\\
Institution1\\
Institution1 address\\
{\tt\small firstauthor@i1.org}
\and
Second Author\\
Institution2\\
First line of institution2 address\\
{\tt\small secondauthor@i2.org}
}

\maketitle
\ificcvfinal\thispagestyle{empty}\fi
%%%%%%%%% Introduction

\section{Details of Proposed Heads}
Given the input feature map, the proposal head predicts a proposal heatmap of shape \(1 \times H_p \times W_p \) and a parameter map of shape \(C_p \times H_p \times W_p\). To balance speed and granularity, we choose the feature map in the resolution of downscale 16 as the input. The detailed architecture of the proposal head is shown in Tabel ~\ref{tab:proposal-head}.
\begin{table}[ht]
\centering
\setlength{\tabcolsep}{4pt}{
\scalebox{0.77}{
\begin{tabular}{l|llll|l}
\hline
layer    &kernel & dim & act & norm & output size \\
\hline
\multicolumn{6}{c}{proposal branch}\\
\hline
\(conv2d_1\) & 3 & 64 & \(relu\) & \(bn\) & \(64\times H_p \times W_p\)\\
\(conv2d_2\) & 3 & 1 & \(sigmoid\) &  & \(1\times H_p \times W_p\)\\
\hline
\multicolumn{6}{c}{parameter branch}\\
\hline
\(conv2d_1\) & 3 & 64 & \(relu\) & \(bn\) & \(64\times H_p \times W_p\)\\
\(conv2d_2\) & 3 & 128\(\,or\,\)134 & \(sigmoid\) & & \((128\,or\,134)\times H_p \times W_p\)\\
\hline
\end{tabular}}}
\caption{The detailed architecture of the proposal head. For the \(conv2d_2\) layer, the output dim is 128 if the proposed RIM is added (For CurveLanes). Else the output dim is 134 (for CULane and TuSimple).}
\label{tab:proposal-head}
\end{table}

The conditional shape head consists of a group of shared layers and two branches of conditional convolutions. We concatenate the absolute coordinates to add the absolute location information to the network~\cite{liu2018intriguing}. Considering that a large resolution contributes to the location accuracy, we choose the feature map with the resolution of a downscale 4 or 8. The detailed architecture of the conditional shape head is shown in Tabel ~\ref{tab:shape-head}.
\begin{table}[ht]
\centering
\setlength{\tabcolsep}{4pt}{
\scalebox{0.85}{
\begin{tabular}{l|llll|l}
\hline
layer    &kernel  & dim & act & norm & output size \\
\hline
\(conv2d_1\) & 3 & 64 & \(relu\) & \(bn\) & \(64\times Y \times X\)\\
\(conv2d_2\) & 3 & 64 & \(relu\) & \(bn\) & \(64\times Y \times X\)\\
\(conv2d_3\) & 3 & 64 &  &  & \(64\times Y \times X\)\\
\(cat\, coord\) & & 66 &  &  & \(66\times Y \times X\)\\
\hline
\multicolumn{6}{c}{location branch}\\
\hline
\(C. conv2d\) & 3 & 1 &  &  & \(1\times Y \times X\)\\
\hline
\multicolumn{6}{c}{offset branch}\\
\hline
\(C.conv2d\) & 3 & 1 & & & \(1\times Y \times X\)\\
\hline
\end{tabular}}}
\caption{The detailed architecture of the conditional shape head.}
\label{tab:shape-head}
\end{table}

\section{More Visualization Results}
{\begin{figure*}[ht]
\centering
\includegraphics[scale=0.48]{CurveLanes.pdf}
\caption{Visualization results of the ablation experiments on CurveLanes. The first column shows the results of the baseline that use the conditional instance segmentation strategy~\cite{tian2020conditional, wang2020solov2}. The second column shows the results of our CondLaneNet, which uses the proposed conditional lane detection strategy with the row-wise formulation(including the offset refinement). The third column adds the transformer encoder. The fourth column adds the proposed RIM. The fifth column is the ground-truth. Different lane instances are represented by different colors.
}
\label{Fig.curve} %用于文内引用的标签
\end{figure*}}

{\begin{figure*}[ht]
\centering
\includegraphics[scale=0.49]{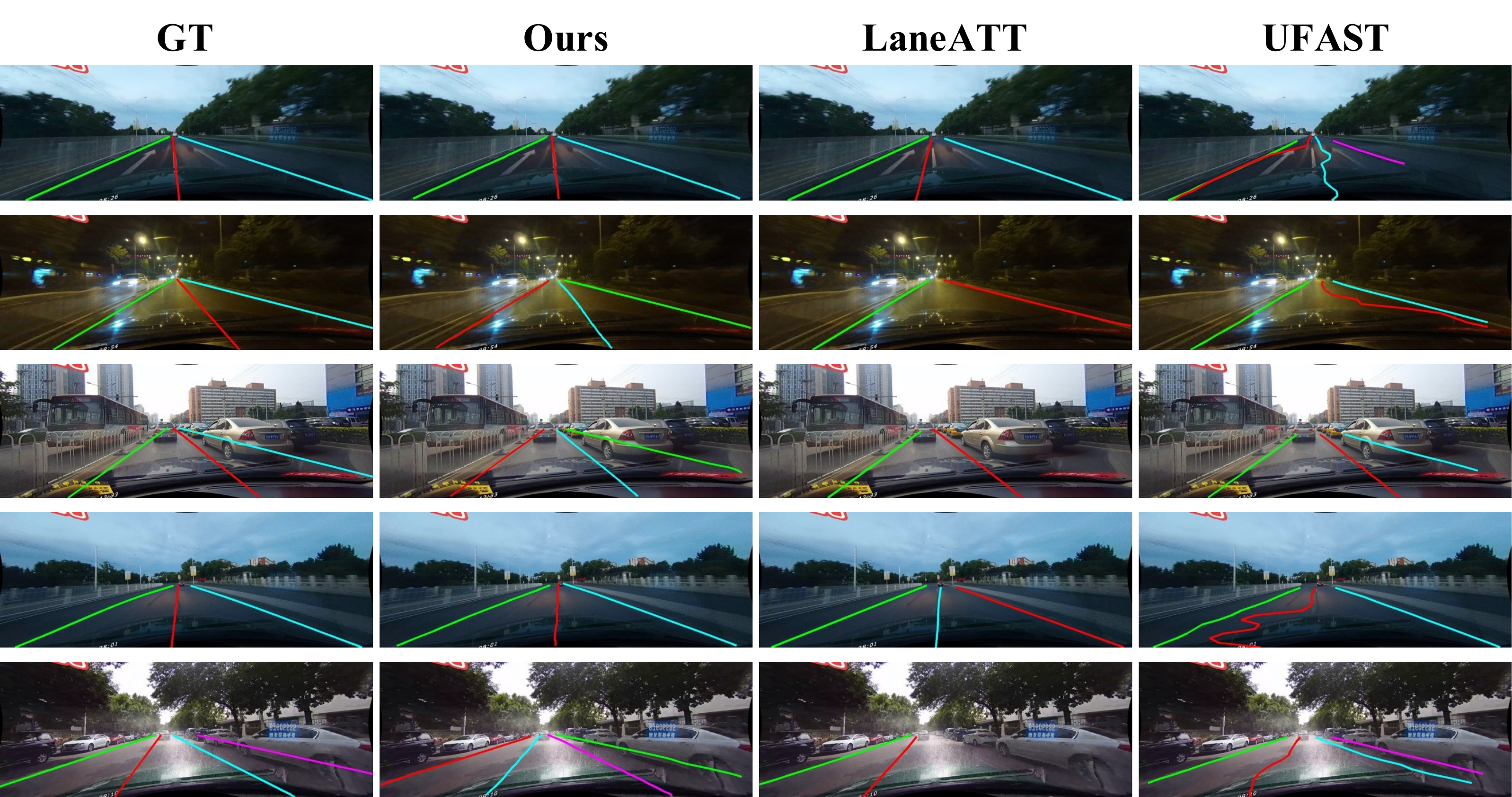}
\caption{Comparison with LaneATT~\cite{tabelini2020keep} and UFAST~\cite{qin2020ultra}. All the results are achieved through the small version with a ResNet-18 backbone~\cite{he2016deep}.
}
\label{Fig.culane} %用于文内引用的标签
\end{figure*}}

\subsection{Results of Ablation Experiments}
To visually show the effect of the proposed improvement strategies, this section provides the visualization results of the ablation experiments on CurveLanes~\cite{li2020curvelane}, as is shown in Figure ~\ref{Fig.curve}. The first column shows the results of the baseline that use the conditional instance segmentation strategy~\cite{tian2020conditional, wang2020solov2}. The second column shows the results of our CondLaneNet, which adopts the proposed conditional lane detection strategy with the row-wise formulation(including the offset post-refinement). The third column adds the transformer encoder. The fourth column adds the proposed RIM. The fifth column is the ground-truth. All the results are based on the small version of our CondLaneNet that utilizes the ResNet-18~\cite{he2016deep} backbone.

The comparison of the first two columns indicates that the proposed conditional lane detection strategy has a stronger instance-level discrimination ability (more lane instances are detected). Moreover, the lines are smoother and have higher local fineness, due to the proposed row-wise formulation with the offset refinement. The transformer encoder mainly improves the detection of the proposal points, especially for the case of no-visual-clue such as the occlusion. Comparing the third and the fourth column, we can observe the function of the proposed RIM, which significantly improves the detection of fork lines and dense lines.

\subsection{Comparison with Other Methods}
We supplement the visualization results of comparison with other state-of-the-art methods on CULane ~\cite{pan2018spatial} in this section, as is shown in Figure ~\ref{Fig.culane}. We compare our results with LaneATT~\cite{tabelini2020keep} (a typical anchor-based method that has state-of-the-art performance on CULane) and UFAST (a typical row-wise detection method) on CULane. The results are all achieve based on the small version with a ResNet-18~\cite{he2016deep} backbone.

Compared with LaneATT~\cite{tabelini2020keep}, in some complex scenarios such as shadow, crowded vehicles, and dazzle light, our method has stronger instance detection performance. 
For UFAST~\cite{qin2020ultra}, the detection of the line in the middle of the image is prone to inaccurate predictions due to the instance-discrimination strategy. Row-wise detection methods adopt the multi-class classification strategy for instance-level discrimination. For CULane, the lane lines are labeled as ``left-left'', ``left'', ``right'', ``right-right'' according to the line position. However, for the line in the middle of the image, this strategy has produced ambiguity and reduces the accuracy of the middle line. In this work, we improve the row-wise formulation with the proposed conditional lane detection strategy and realize more flexible instance-level discrimination.

{\small
\bibliographystyle{ieee_fullname}
\bibliography{egbib}
}